\documentclass[conference]{IEEEtran}
\IEEEoverridecommandlockouts

\usepackage{amsmath,amssymb,amsfonts}
\usepackage{algorithmic}
\usepackage{graphicx}
\usepackage{textcomp}
\usepackage{xcolor}
\usepackage{hyperref} 
\usepackage{listings}
\usepackage{xcolor}
\usepackage[most]{tcolorbox}
\usepackage{subcaption}

\usepackage{adjustbox}
\usepackage{booktabs} 
\usepackage[T1]{fontenc}
\usepackage[utf8]{inputenc}

\def\BibTeX{{\rm B\kern-.05em{\sc i\kern-.025em b}\kern-.08em
    T\kern-.1667em\lower.7ex\hbox{E}\kern-.125emX}}
\begin{document}

\title{AraFinNews: Arabic Financial Summarisation with Domain-Adapted LLMs}

\author{\IEEEauthorblockN{1\textsuperscript{st}Mo El-Haj}
\IEEEauthorblockA{\textit{NLP @ VinUni --- UCREL NLP} \\
\textit{VinUniversity --- Lancaster University}\\
Hanoi, Vietnam --- Lancaster, UK \\
elhaj.m@vinuni.edu.vn --- m.el-haj@lancaster.ac.uk}
\and
\IEEEauthorblockN{2\textsuperscript{nd} Paul Rayson}
\IEEEauthorblockA{\textit{School of Computing and Communications} \\
\textit{Lancaster University}\\
Lancaster, UK \\
p.rayson@lancaster.ac.uk}
}

\maketitle

\begin{abstract}
We introduce \textit{AraFinNews}, the largest publicly available Arabic financial news dataset to date, comprising 212{,}500 article–headline pairs spanning a decade of reporting from 2015 to 2025. Designed as an Arabic counterpart to major English summarisation corpora such as CNN/DailyMail, AraFinNews provides a realistic benchmark for evaluating domain-specific language understanding and generation in financial contexts. Using this resource, we investigate the impact of domain specificity on abstractive summarisation of Arabic financial texts with large language models (LLMs). In particular, we evaluate transformer-based models—mT5, AraT5, and the domain-adapted FinAraT5—to examine how financial-domain pretraining influences accuracy, numerical reliability, and stylistic alignment with professional reporting. Experimental results show that domain-adapted models generate more coherent summaries, especially in their handling of quantitative and entity-centric information. These findings highlight the importance of domain-specific adaptation for improving narrative fluency in Arabic financial summarisation. The dataset is freely available for non-commercial research at \url{https://github.com/ArabicNLP-uk/AraFinNews}.
\end{abstract}


\section{Introduction}
\label{sec:introduction}

Arabic, spoken by more than 300 million people across over twenty countries~\cite{statista2024}, plays a central role in religious, political, and cultural communication throughout the Arab world. Yet, despite its global significance, the language remains underrepresented in Natural Language Processing (NLP). Efforts such as ACL SIGARAB\footnote{\url{https://www.sigarab.org/}} have increased visibility and coordination across the community, but the development of robust, domain-specific Arabic language models and datasets continues to lag behind progress in high-resource languages such as English and Chinese~\cite{el2010using}.

This gap is driven by persistent data scarcity and the linguistic characteristics of Arabic itself. The language is diglossic, spanning Classical Arabic, Modern Standard Arabic (MSA), and a wide range of regional dialects that vary lexically, morphologically, and syntactically. In addition, its right-to-left script, lack of capitalisation, and frequent omission of diacritics—short vowel markers that help disambiguate homographs—create further difficulties for tokenisation, parsing, and semantic interpretation~\cite{el-haj-rayson-2016-osman}.

Recent advances in Arabic NLP—through models such as AraBERT~\cite{antoun_arabert_2021}, ARBERT, MARBERT~\cite{abdul-mageed_arbert_2021}, and generative models like AraBART and AraT5~\cite{eddine_arabart_2022, nagoudi_arat5_2022}—have substantially improved general text understanding. However, these models are typically trained on broad web or news data and therefore lack consistent exposure to specialised financial or technical language. By contrast, English NLP has benefited from domain-adapted architectures such as FinBERT~\cite{yang2020finbert} and BioBERT~\cite{lee2020biobert}, which have demonstrated clear gains in financial and biomedical tasks. The absence of similar domain adaptation efforts for Arabic underscores the need for targeted pretraining strategies capable of capturing the linguistic and conceptual characteristics of financial discourse.

This need is amplified by the rapid expansion of the Arab financial sector over the past two decades, with markets such as Tadawul\footnote{\url{https://www.saudiexchange.sa}}, the Dubai Financial Market (DFM)\footnote{\url{https://www.dfm.ae}}, and the Abu Dhabi Securities Exchange (ADX)\footnote{\url{https://www.adx.ae}} now ranking among the world’s most active~\cite{zmandar2021joint}. The resulting increase in Arabic financial narratives—corporate disclosures, regulatory filings, market analyses, and news reports—has produced texts that are long, information-dense, and stylistically formal, making them well suited to automatic summarisation~\cite{zmandar2021multilingual}. Yet most Arabic summarisation systems continue to focus on general news, leaving financial text underexplored despite its demand for accuracy, numerical precision, and stylistic clarity~\cite{el2020financial,zmandar2021financial,zavitsanos2023financial}.

To address this gap, we introduce \textbf{AraFinNews}—the largest publicly available Arabic financial news dataset, comprising 212{,}500 article–headline pairs from reputable financial media sources. Designed as the Arabic equivalent of major English summarisation benchmarks such as CNN/DailyMail, AraFinNews provides a rich domain-specific resource for abstractive summarisation and related financial NLP tasks. It is freely available for research at \url{https://github.com/ArabicNLP-UK/AraFinNews}. Using this dataset, we evaluate transformer-based architectures—including FinAraT5, AraT5, and mT5—on Arabic financial summarisation. This work contributes to advancing domain-specific Arabic NLP, demonstrating the benefits and challenges of adapting large language models to one of the world’s most linguistically complex and economically dynamic languages.

\section{Related Work}
\label{sec:relatedwork}

Text summarisation has long been a core task in Natural Language Processing (NLP), particularly within journalism and information retrieval. Early systems were largely extractive, relying on sentence ranking and statistical heuristics such as term frequency, cue words, and sentence position~\cite{filippova2009company, berger2000ocelot}. Although these methods captured surface-level salience, they struggled to produce coherent, concise outputs and often generated redundant summaries.

The shift to neural encoder–decoder architectures marked a turning point. Recurrent models with attention mechanisms enabled systems to generate novel text rather than simply selecting existing sentences. One of the earliest demonstrations of fluent headline generation using LSTM-based architectures was presented in~\cite{lopyrev2015generating}. Subsequent improvements incorporated coverage mechanisms and reinforcement learning to reduce redundancy and improve factual consistency~\cite{see-etal-2017-get, cheng-etal-2016-neural}.

Transformer-based models such as BART and T5 further advanced abstractive summarisation. Their text-to-text formulation supported strong generalisation across languages and tasks~\cite{raffel_exploring_2020}, while multilingual variants like mT5~\cite{xue-etal-2021-mt5} extended these capabilities to low-resource settings through cross-lingual transfer. However, these broadly pretrained models often struggle in specialised domains—such as finance—where numerical reasoning, formal style, and domain-specific terminology require more targeted adaptation.

Recent developments in Arabic NLP have mirrored these global trends. Large-scale pretrained models such as AraBERT~\cite{antoun_arabert_2021}, ARBERT and MARBERT~\cite{abdul-mageed_arbert_2021} have become standard baselines for classification and sentiment analysis, addressing challenges related to rich morphology and dialectal variation. Generative models including AraBART and AraT5~\cite{eddine_arabart_2022, nagoudi_arat5_2022}, and more recent efforts such as Jasmine~\cite{abdul2023jasmine}, have extended these capabilities to summarisation and text generation tasks.

A growing body of research now focuses specifically on Arabic abstractive summarisation. \cite{kahla2023fine} demonstrated the benefits of fine-tuning multilingual transformers for Arabic news summarisation, while \cite{morsi2025arabic} proposed transformer-based architectures that improved coherence and fluency over earlier systems. Other work has integrated linguistic information or auxiliary objectives- for example, \cite{essa2025enhanced} combined named entity recognition with summarisation to improve factual preservation, and \cite{elsaid2025hybrid} developed a hybrid RNN–Transformer model to enhance grammatical consistency. Collectively, these studies highlight the feasibility of generative approaches for Arabic, though most efforts remain centred on general news rather than specialised professional domains.

Despite the progress, current Arabic language models are still predominantly general-purpose. Models such as AraBERT, ARBERT, MARBERT, AraBART, and AraT5 are trained on broad web or news corpora and therefore have limited exposure to financial or technical text. A notable exception is FinAraT5~\cite{zmandar2023finarat5}, a domain-adapted variant further pretrained on large Arabic financial corpora. However, such efforts remain rare compared with English NLP, where models like FinBERT~\cite{yang2020finbert} and BioBERT~\cite{lee2020biobert} demonstrate the effectiveness of targeted pretraining for specialised tasks. The scarcity of comparable domain-specific resources for Arabic reinforces the need for focused pretraining and systematic evaluation in financial summarisation.

\subsection{Datasets for Arabic Summarisation and Financial Corpora}

Research in Arabic summarisation has been supported by a range of corpora covering both extractive and abstractive paradigms. Early resources such as Kalimat~\cite{ElHaj2013KALIMATAM}, a 20{,}000-article Omani newspaper corpus, and the Essex Arabic Summaries Corpus (EASC)~\cite{el2010using}, which includes 153 news articles and 765 human-written summaries, provided the foundation for extractive approaches. More recent multilingual datasets have enabled progress in neural and cross-lingual methods. The Arabic Gigaword corpus~\cite{Parker2019gigawords}, although large and widely used, lacks aligned document–summary pairs and remains commercially restricted. In contrast, openly available abstractive benchmarks such as XLSum~\cite{hasan-etal-2021-xl} and WikiLingua~\cite{ladhak-etal-2020-wikilingua} introduced approximately 40{,}000 and 21{,}000 Arabic examples respectively, though their content diverges from financial and professional prose. 

Additional Arabic-specific datasets have further expanded training resources. SumArabic~\cite{SumArabic} provides 84{,}764 high-quality abstractive text–summary pairs built from Common Crawl data and has been used to benchmark RNN-based and Transformer-based models. HASD~\cite{HASD} introduces a 43{,}000-article benchmark containing both extractive and abstractive summaries, alongside an additional 150{,}000-sample abstractive dataset (AASD) and a modified version of EASC with new abstractive annotations. While these datasets have accelerated progress in Arabic abstractive summarisation, they target general content and do not capture the stylistic and numerical characteristics of financial narratives.

Earlier work on Arabic summarisation relied heavily on extractive methods such as TF–IDF ranking, lexical scoring, and graph-based centrality measures~\cite{El-Haj-2011, Douzidia2004LakhasAA, el2013using, koulali2013arabic}. More recent studies have shifted towards neural models, with multilingual architectures such as mBART and mT5 fine-tuned for Arabic summarisation~\cite{kahla-etal-2021-cross}, yielding improvements in fluency and abstraction.

However, unlike English, Arabic still lacks large, curated corpora for specialised domains. Resources such as BORSAH~\cite{ALSHAHRANI18} and the Arabic Business and Management Corpus (ABMC)\footnote{\url{https://elhaj.uk/corpora.html}} offer valuable financial and economic text collections, but their primary use has been sentiment analysis and classification rather than summarisation. As a result, progress in Arabic abstractive summarisation—particularly for financial text—has been limited by the absence of domain-focused, large-scale datasets.

Despite this progress, existing resources still fail to meet the requirements of Arabic financial summarisation. Most available datasets focus on general news or multilingual content, leaving domain-specific financial material largely unaddressed. The only notable effort in this direction is the FinAraSum dataset~\cite{zmandar2023finarat5}, which contains approximately 44{,}900 financial news articles collected from the Al Arabiya Aswaq website over a decade and was designed for headline generation. While valuable, FinAraSum is limited in size, covers a narrower range of financial reporting, and—crucially—has not been released publicly, restricting reproducibility and wider use. As a result, researchers lack an open, large-scale Arabic financial summarisation benchmark, leaving a significant gap in resources for training, evaluation, and domain-adapted modelling.

The \textbf{AraFinNews} dataset introduced in this paper fills this gap by providing over 200{,}000 high-quality Arabic financial article–headline pairs, enabling research in summarisation and broader financial NLP.

This unmet intersection between Arabic summarisation and financial text processing presents a significant research opportunity: building datasets and models that can handle numerically dense, terminologically specialised, and stylistically formal financial narratives. Addressing this need is essential for developing practical, context-aware summarisation systems for professional Arabic applications.

\section{AraFinNews: The Arabic Financial News Dataset}
\label{sec:dataset}

To address the lack of large, domain-specific resources for Arabic summarisation, we introduce AraFinNews, the largest publicly available Arabic financial news dataset to date. The corpus comprises 212,500 article–headline pairs sourced through Argaam.com, a major financial media platform in the Gulf region that aggregates material from several well-established outlets, including world financial news from the Financial Times, Forbes, the World Gold Council and Bloomberg. Each headline serves as an abstractive summary of its corresponding article, reflecting the concise, information-rich style characteristic of professional Arabic financial journalism. The dataset spans corporate disclosures, market analyses and macroeconomic reports, offering a solid benchmark for abstractive summarisation as well as related NLP tasks such as named entity recognition, event extraction and sentiment analysis.

AraFinNews covers a continuous period of nearly ten years, from 8 October 2015 to 29 July 2025, capturing key economic cycles, market events, and policy developments across the world. This long-term temporal span ensures the dataset reflects evolving financial language, trends, and reporting styles over time—making it a valuable resource for both synchronic and diachronic analyses of Arabic financial discourse.

In scope and structure, AraFinNews is comparable to influential English summarisation corpora such as CNN/DailyMail, serving as a domain-specific benchmark for evaluating the effectiveness of Arabic large language models in financial text understanding and generation. The dataset is freely accessible for non-commercial research\footnote{\url{https://github.com/ArabicNLP-UK/AraFinNews}}.

All content was collected in accordance with Argaam’s publicly accessible content policy and complies with ethical data usage standards. Argaam’s \texttt{robots.txt} file does not prohibit web scraping, and all retrieved material was publicly available without subscription or authentication. The released dataset includes comprehensive metadata fields—article ID, title, date, full text, and source URL—to facilitate transparent and reproducible research.

\subsection{Corpus Composition}

AraFinNews contains 212{,}512 Arabic financial news articles, each paired with its original headline. The corpus is released as a large CSV file and every record is stored in JSON format with the following fields:

\begin{itemize}
    \item \texttt{id}: unique numeric identifier.
    \item \texttt{title}: headline text (serving as a short abstractive summary).
    \item \texttt{date}: publication date.
    \item \texttt{article}: full article body.
\end{itemize}

This structure ensures seamless integration with widely used machine learning frameworks such as HuggingFace Datasets and PyTorch. The corpus retains Modern Standard Arabic (MSA) text in its natural form—without diacritics or orthographic normalisation—to preserve the authentic characteristics of professional financial reporting.

\subsection{AraFinNews Corpus Statistics}

Table~\ref{tab:financial_stats} presents the main linguistic and structural statistics of AraFinNews. The dataset exhibits a rich lexical distribution and a high density of numerical and named-entity expressions, reflecting the precision and formal tone typical of Arabic financial journalism.

\begin{table}[htbp]
\centering
\resizebox{\columnwidth}{!}{%
\begin{tabular}{lrrr}
\toprule
\textbf{Metric} & \textbf{Titles} & \textbf{Articles} & \textbf{Combined} \\
\midrule
Total words & 2{,}384{,}000 & 39{,}951{,}000 & 42{,}335{,}000 \\
Unique vocabulary & 69{,}900 & 542{,}300 & 547{,}000 \\
Numerical tokens (\%) & 4.2\% & 5.3\% & 5.3\% \\
Average word length (chars) & 4.9 & 4.8 & 4.8 \\
Average title length (words) & 11.2 & --- & --- \\
Average article length (words) & --- & 188 & --- \\
\bottomrule
\end{tabular}
}
\caption{Descriptive statistics for the AraFinNews dataset.}
\label{tab:financial_stats}
\end{table}

\subsection{Title-to-Article Ratio}

In AraFinNews, titles represent on average around 6\% of the length of their corresponding articles—roughly one title word for every sixteen article words. The median ratio is about 5.5\%, reflecting natural variation in conciseness across the corpus. Articles contain an average of 188 tokens, while titles average 11 tokens, with approximately 5\% of all tokens being numerical. This quantitative density underscores the dataset’s suitability for financial NLP applications that require precise handling of figures, entities, and measurements.

\subsection{Data Processing and Cleaning}
\label{ssec:data-cleaning}

Although Argaam articles are generally well-structured, standardisation was required to prepare them for research use. Each HTML page was parsed to extract the \texttt{title}, \texttt{date}, and \texttt{article} fields before storing them in JSON format. Preprocessing involved:
\begin{itemize}
    \item Removing HTML tags, residual markup, extraneous characters, links, and images.
    \item Normalising date formats.
    \item Filtering duplicate entries caused by syndication or reposting.
\end{itemize}

This lightweight, field-level cleaning preserved the original editorial form of the Modern Standard Arabic content: no diacritics, tatweel (elongation) characters, or orthographic variants were altered, and no additional token-level normalisation was applied.

\subsection{Data Splits}
\label{ssec:data-splits}

For benchmarking and reproducibility, the dataset was divided into training, validation, and test splits using an 80/10/10 ratio, ensuring temporal and topical diversity across all partitions. The split was performed using the unique \texttt{id} field of each article to guarantee deterministic reproducibility, with the resulting index files released as \texttt{AraFinNews\_train\_ids.csv}, \texttt{AraFinNews\_val\_ids.csv}, and \texttt{AraFinNews\_test\_ids.csv}. While this stratified random sampling approach maintains balanced coverage across sectors and publication periods, it may allow minor temporal overlap between training and test sets. As the FinAraSum dataset, on which our setup was modelled, is not publicly available, we were unable to replicate its original split protocol precisely. 

The final distribution comprises \textbf{170,012 training}, \textbf{21,251 validation}, and \textbf{21,252 test} articles (Table~\ref{tab:dataset_stats_comparison}). This structure aligns with common summarisation datasets such as CNN/DailyMail and FinAraSum, facilitating fair cross-dataset comparison and enabling model evaluation under consistent conditions.

\subsection{Vocabulary and Tokenisation}

For tokenisation, AraFinNews employs the \texttt{SentencePiece} tokenizer from AraT5, trained with a unigram language model comprising 110,000 subword units optimised for Modern Standard Arabic. This configuration provides broad lexical coverage across proper nouns, financial terminology, and numerical expressions, reducing out-of-vocabulary rates and ensuring direct compatibility with existing Arabic generative language models.

During model fine-tuning, each architecture used its native tokenisation scheme to preserve compatibility with its pretrained vocabulary: AraT5 and FinAraT5 employed the AraT5 SentencePiece model (110k tokens), mT5 variants used the original multilingual SentencePiece vocabulary (250k tokens), and BERT2BERT relied on the standard mBERT WordPiece tokenizer.

\subsection{Sample Entry}

\begin{figure}[htbp]
\centering
\includegraphics[width=\linewidth]{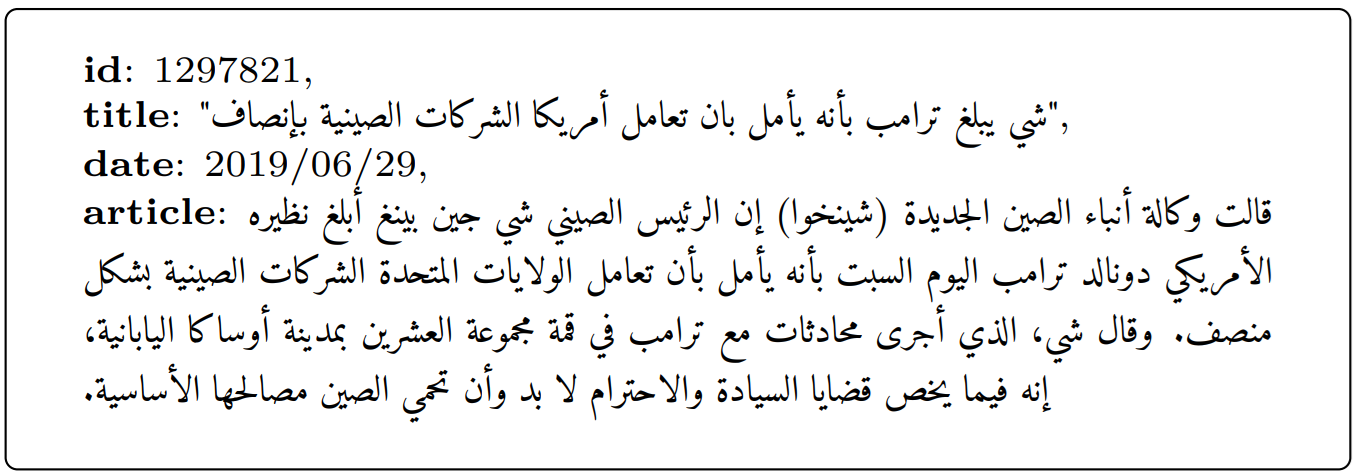}
\caption{Arabic example record from AraFinNews}
\label{fig:arabic_sample}
\end{figure}

\begin{figure}[htbp]
\centering
\includegraphics[width=\linewidth]{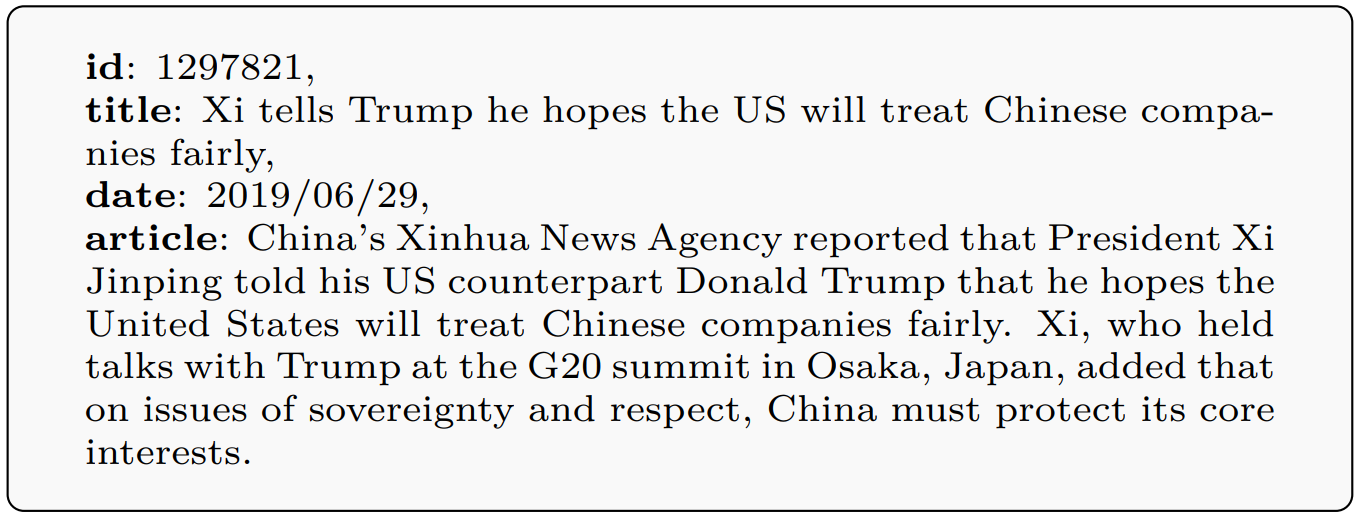}
\caption{English translation of the Arabic example record from AraFinNews}
\label{fig:english_sample}
\end{figure}

An example record from AraFinNews is shown in Figure \ref{fig:arabic_sample}, followed by its English translation in Figure \ref{fig:english_sample}.

This consistent structure makes AraFinNews suitable for large-scale model training and evaluation in Arabic abstractive summarisation, while also serving as a strong foundation for domain-specific research in financial NLP.

\subsection{Dataset Comparison and Abstractiveness Analysis}
\label{ssec:dataset_comparison}

To better contextualise the characteristics of our dataset, we compare the \textbf{AraFinNews} corpus with well-established English summarisation datasets, including CNN, DailyMail, and the Arabic financial summarisation benchmark \textbf{FinAraSum} \cite{zmandar2023finarat5}. Table~\ref{tab:dataset_stats_comparison} summarises the basic statistics, while Table~\ref{tab:dataset_abstractiveness_comparison} reports abstractiveness and baseline performance using the LEAD-1 metric.

AraFinNews is comparable in nature to FinAraSum, both being derived from financial newswire content where the article body and headline reflect domain-specific reporting. However, AraFinNews contains over 210K articles—substantially larger than FinAraSum (44.9K).

The average article length in AraFinNews is approximately 188, while the average headline length is \textbf{11 words}, closely mirroring FinAraSum (238.3 and 9 words, respectively). Compared with English summarisation corpora such as CNN and DailyMail, the Arabic datasets contain significantly shorter summaries, reflecting the highly condensed nature of Arabic financial headlines that often consist of a single clause or nominal phrase.

Table~\ref{tab:dataset_abstractiveness_comparison} shows that AraFinNews is a highly \textbf{abstractive} dataset, with \textbf{35.6\% novel unigrams} and \textbf{over 91\% novel 4-grams} in gold summaries—very similar to FinAraSum (37.8\% unigrams and 95.2\% 4-grams). This suggests that models trained on AraFinNews must generate original, domain-grounded content rather than relying on extractive copying. Compared to CNN and DailyMail (which have unigram novelty below 18\%), AraFinNews presents a substantially more challenging summarisation task.

Despite their brevity, AraFinNews headlines yield moderate LEAD-1 scores (ROUGE-1 = 11.6, ROUGE-2 = 1.1, ROUGE-L = 11.6). These results reflect the dataset’s highly abstractive nature and the linguistic complexity of Arabic, underscoring AraFinNews as a challenging and realistic benchmark for evaluating Arabic financial text generation.

These statistics confirm that AraFinNews shares the core properties of FinAraSum abstractiveness—while offering broader coverage and scale. Its characteristics make it a valuable resource for benchmarking Arabic headline generation and for evaluating domain-specific language models such as FinAraT5.

\begin{table}[htbp]
\centering
\begin{adjustbox}{width=1\columnwidth}
\begin{tabular}{|l|c|c|c|c|c|}
\hline
\textbf{Dataset} & \textbf{Train / Val / Test} & \textbf{Avg. Doc. Len.} & \textbf{Avg. Sum. Len.} & \textbf{Vocab (Docs)} & \textbf{Vocab (Sums)} \\
\hline
CNN & 90{,}266 / 1{,}220 / 1{,}093 & 760.5 & 45.7 & 340K & 89K \\
DailyMail & 197{,}461 / 12{,}148 / 10{,}397 & 653.3 & 54.6 & 564K & 180K \\
FinAraSum & 44{,}900 / 2{,}800 / 2{,}800 & 238.3 & 9.0 & 493K & 79K \\
\textbf{AraFinNews (ours)} & \textbf{170{,}000 / 21{,}250 / 21{,}250} & \textbf{188} & \textbf{11.22} & \textbf{528K} & \textbf{83K} \\
\hline
\end{tabular}
\end{adjustbox}
\caption{Comparison of dataset statistics between AraFinNews and existing summarisation datasets. Train/validation/test sizes are given in number of documents. Average document and summary lengths are in words. Vocabulary sizes (unique tokens) are shown in thousands. Figures for English datasets are based on standard splits reported in prior work~\cite{kamal-eddine-etal-2021-barthez}.}
\label{tab:dataset_stats_comparison}
\end{table}

\begin{table}[htbp]
\centering
\begin{adjustbox}{width=1\columnwidth}
\begin{tabular}{|l|c|c|c|c|c|c|c|}
\hline
\textbf{Dataset} & \textbf{Unigrams} & \textbf{Bigrams} & \textbf{Trigrams} & \textbf{4-grams} & \textbf{R1} & \textbf{R2} & \textbf{RL} \\
\hline
CNN & 16.8 & 54.3 & 72.4 & 80.4 & 29.2 & 11.1 & 25.9 \\
DailyMail & 17.0 & 53.8 & 72.1 & 80.3 & 40.7 & 18.4 & 37.3 \\
FinAraSum & 37.8 & 73.6 & 89.0 & 95.2 & 18.3 & 7.1 & 14.8 \\
\textbf{AraFinNews (ours)} & \textbf{35.6} & \textbf{67.8} & \textbf{83.5} & \textbf{91.3} & \textbf{11.6} & \textbf{1.1} & \textbf{11.6} \\
\hline
\end{tabular}
\end{adjustbox}
\caption{Comparison of abstractiveness and LEAD-1 ROUGE scores across datasets.}
\label{tab:dataset_abstractiveness_comparison}
\end{table}

\section{Financial Large Language Models for Arabic Summarisation}
\label{sec:financial_llms}

Building on the AraFinNews dataset introduced in Section~\ref{sec:dataset}, this section explores how large language models (LLMs) can leverage large-scale, domain-specific data to advance abstractive summarisation in Arabic financial journalism.

Recent advancements in LLMs have transformed financial text processing, enabling sophisticated applications such as document summarisation, sentiment analysis, and report generation. Prominent models such as GPT~\cite{brown2020language}, PaLM~\cite{chowdhery2022palm}, and FLAN-T5~\cite{chung2022scaling} have achieved state-of-the-art performance in English and multilingual settings. However, their effectiveness in Arabic remains limited by the scarcity of large, high-quality domain-specific corpora and by the inherent linguistic challenges of the language—particularly its morphological richness, diglossic variation, and orthographic ambiguity.

To address the limitations of general-purpose Arabic models, recent work introduced Arabic-adapted transformers such as AraBERT~\cite{antoun2020arabert} and AraT5~\cite{nagoudi_arat5_2022}, as well as the domain-specialised FinAraT5~\cite{zmandar2023finarat5}\footnote{\url{https://huggingface.co/drelhaj/FinAraT5}}. In this study, we evaluate these models—alongside multilingual baselines—under a unified experimental setup to assess their ability to generate coherent financial headlines for AraFinNews. The following subsections detail the model selection, fine-tuning configuration, and evaluation framework adopted in this work.

\subsection{Model Selection and Comparative Setup}

To benchmark Arabic financial summarisation effectively, we implemented several encoder–decoder transformer architectures within a unified text-to-text framework (Table \ref{tab:llm_models}). FinAraT5 was selected as the primary model due to its continued pretraining on large-scale Arabic financial documents, enabling it to capture domain-specific terminology, numerical reasoning, and stylistic conventions characteristic of financial reporting. For comparative evaluation, we included mT5~\cite{xue-etal-2021-mt5} as a multilingual baseline and AraT5~\cite{nagoudi_arat5_2022} as a general-purpose Arabic model. Together, these systems represent three distinct pretraining paradigms: multilingual generalisation, monolingual adaptation, and financial-domain specialisation.

\begin{table}[htbp]
\centering
\resizebox{\columnwidth}{!}{%
\begin{tabular}{lll}
\hline
\textbf{Model} & \textbf{Domain} & \textbf{Architecture} \\
\hline
mT5-Base~\cite{xue-etal-2021-mt5} & Multilingual & Encoder–Decoder (220M) \\
AraT5-Base~\cite{nagoudi_arat5_2022} & General Arabic & Encoder–Decoder (220M) \\
FinAraT5-Base~\cite{zmandar2023finarat5} & Arabic Financial & Encoder–Decoder (220M) \\
\hline
\end{tabular}
}
\caption{Models used in Arabic financial summarisation experiments.}
\label{tab:llm_models}
\end{table}

All models adopt a text-to-text paradigm supporting multiple generation tasks through consistent input–output formatting. Each model was fine-tuned on the AraFinNews training split using identical hyperparameters to ensure comparability, with the development set used for validation and the held-out test set reserved exclusively for final evaluation. Fine-tuning employed the Hugging Face \texttt{Transformers} library with the AdamW optimiser, a learning rate of $5\times10^{-5}$ with linear warm-up, an effective batch size of 32 achieved via gradient accumulation over four steps, and a maximum sequence length of 512 tokens. Evaluation assesses how domain adaptation influences summarisation performance.

\subsection{Arabic Financial Summarisation Pipeline}

Our summarisation pipeline integrates FinAraT5~\cite{zmandar2023finarat5} as the core generation component for producing abstractive financial headlines. The model operates within a sequence-to-sequence framework based on AraT5~\cite{nagoudi_arat5_2022}, following the text-to-text paradigm of T5~\cite{raffel_exploring_2020}. Within this setup, full-length financial news articles are transformed into concise headline-style summaries that reflect the conventions of Modern Standard Arabic financial reporting.

\begin{figure}[htbp]
\centering
\includegraphics[width=\linewidth]{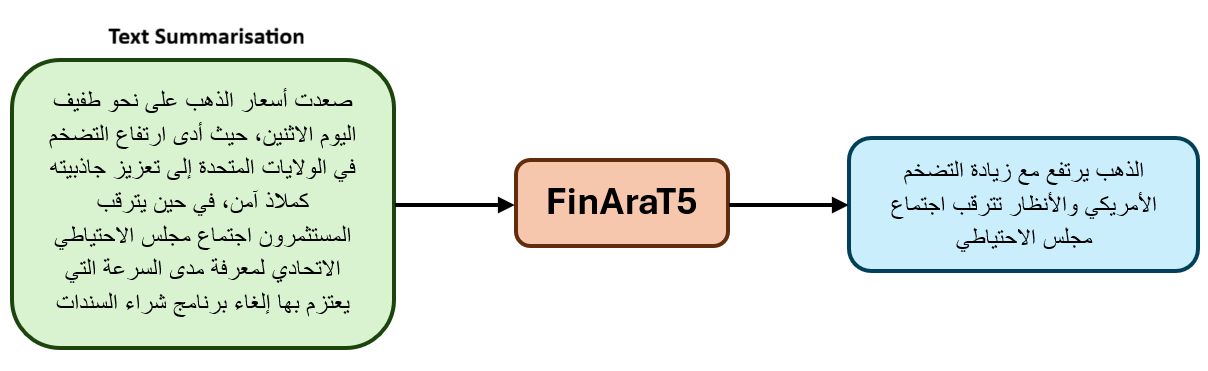}
\caption{Arabic financial headline generation using FinAraT5.}
\label{fig:FinAraT5_pipeline}
\end{figure}

Figure~\ref{fig:FinAraT5_pipeline} illustrates the overall workflow, in which FinAraT5 receives the processed article text as input and generates an abstractive headline output. This component forms the central step of the pipeline used to evaluate summarisation quality on AraFinNews.

FinAraT5 was pretrained using the span-corruption denoising objective introduced in the original T5 framework~\cite{raffel_exploring_2020}, where contiguous spans of text are randomly masked and the model learns to reconstruct the missing content. This self-supervised objective enables large-scale learning without labelled data and captures long-range contextual dependencies.

\begin{figure}[htbp]
\centering
\includegraphics[width=\columnwidth]{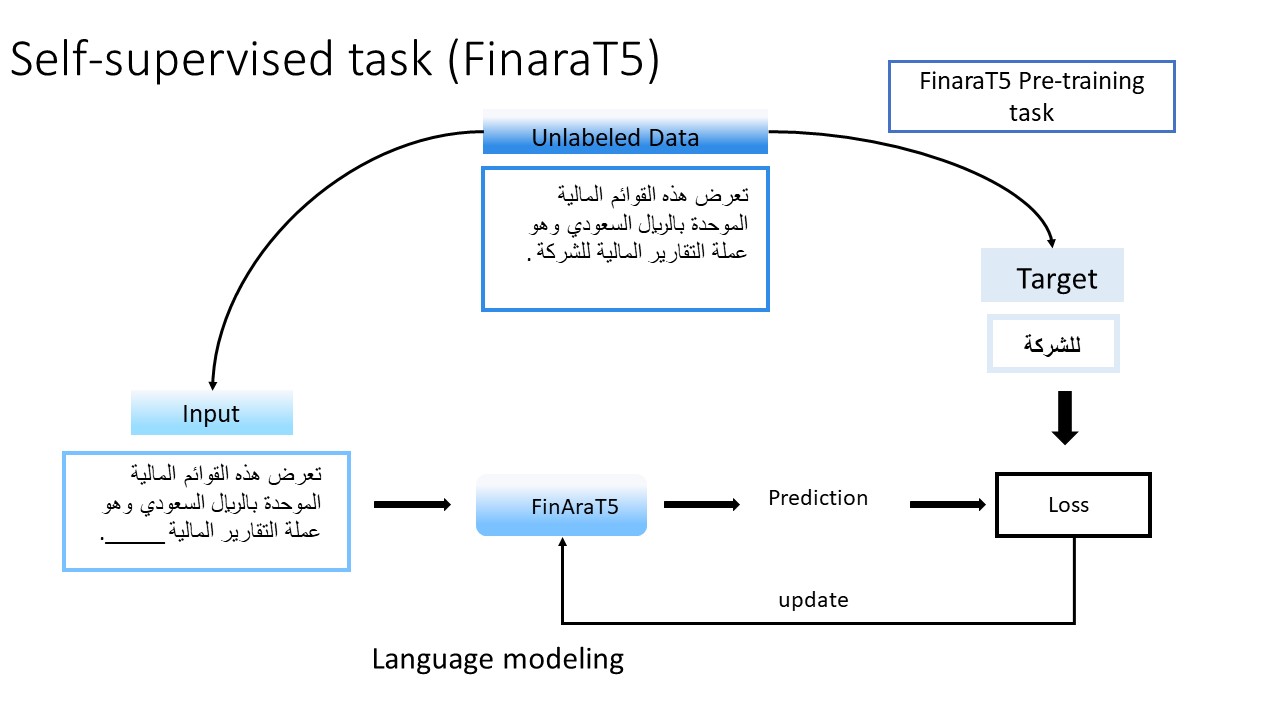}
\caption{Span-corruption pretraining task used in FinAraT5.}
\label{fig:FinAraT5_pretraining_task}
\end{figure}

Building on the methodology of AraT5~\cite{nagoudi_arat5_2022}, FinAraT5 extends this pretraining through continued exposure to raw Arabic financial documents, including large-scale news sources such as AraFinNews. This additional phase enriches the model’s handling of financial terminology, numerical expressions, and the discourse structures characteristic of economic reporting. Figure~\ref{fig:FinAraT5_pretraining_task} illustrates the span-corruption task used during pretraining.

FinAraT5 was trained using the original TensorFlow implementation of T5\footnote{\url{https://github.com/google-research/text-to-text-transfer-transformer}} and managed via the \texttt{t5} library\footnote{\url{https://pypi.org/project/t5/}}. As the model predates later frameworks such as T5X, distributed training was performed using Mesh TensorFlow.

This pretraining strategy provides a strong foundation for downstream generative tasks, particularly abstractive summarisation. Continued exposure to financial text improves the model’s treatment of specialised vocabulary and high-frequency expressions such as (“stocks, profits, indices, returns”), offering advantages over general-purpose Arabic models like AraT5 and mT5. The fine-tuning configuration used in this study is summarised in Table~\ref{tab:training_config}.

\begin{table}[htbp]
\centering
\resizebox{\columnwidth}{!}{%
\begin{tabular}{ll}
\hline
\textbf{Parameter} & \textbf{Value} \\
\hline
Learning rate & $5\times10^{-5}$ (with warm-up) \\
Batch size & 8 (gradient accumulation = 4) \\
Effective batch size & 32 \\
Max sequence length & 512 tokens \\
Epochs & 22 \\
Optimiser & AdamW \\
Loss function & Cross-Entropy \\
Evaluation metrics & ROUGE-1/2/L, METEOR, BARTScore, BERTScore, FrugalScore \\
\hline
\end{tabular}
}
\caption{Fine-tuning configuration for Arabic financial summarisation.}
\label{tab:training_config}
\end{table}

\section{Experiments}
\label{sec:experiments}

\subsection{Arabic Financial Headline Generation}

This section presents experiments on the automatic generation of concise and informative Arabic financial headlines from full-length news articles in the AraFinNews dataset. The task represents a challenging form of abstractive summarisation requiring precision and coherence, and stylistic adherence to the conventions of Modern Standard Arabic (MSA) journalism. Unlike extractive summarisation, which selects sentences directly from the source, abstractive summarisation demands deep semantic interpretation and the generation of novel, well-formed text. In the financial domain, this challenge is heightened by the prevalence of numerical data, company names, and temporal expressions.

We evaluate several pretrained transformer architectures—monolingual, multilingual, and domain-adapted—to assess how domain-specific pretraining affects Arabic financial summarisation. Our focus is on FinAraT5~\cite{zmandar2023finarat5}, a Financial Arabic Large Language Model derived from AraT5 via continued pretraining on large-scale Arabic financial corpora, including data comparable in style and scope to AraFinNews.

\subsubsection*{Model Setup and Baselines}

All models were fine-tuned on the AraFinNews training split using identical training settings for fair comparison. Specifically, we fine-tuned:
\begin{itemize}
    \item \textbf{AraT5} (small, base)~\cite{nagoudi_arat5_2022}: a monolingual Arabic text-to-text transformer trained on general-domain corpora.
    \item \textbf{mT5} (small, base, large)~\cite{xue-etal-2021-mt5}: a multilingual encoder–decoder model covering 101 languages, serving as a strong cross-lingual baseline.
    \item \textbf{FinAraT5} (base): a domain-adapted Arabic T5 variant pretrained on Modern Standard Arabic financial text.
    \item \textbf{BERT2BERT}~\cite{rothe-etal-2020-leveraging}: a multilingual encoder–decoder model using mBERT for both encoding and decoding.
    \item \textbf{Lead-1}: an extractive lower-bound baseline that selects the first sentence of each article as the summary.
\end{itemize}

Each model was trained for up to 22 epochs with a learning rate warmed up to $5\times10^{-5}$, a batch size of 8, and gradient accumulation over 4 steps (yielding an effective batch of 32). Early stopping was applied based on validation loss. All experiments were conducted on Lancaster University’s High-End Computing (HEC) cluster\footnote{\url{https://www.lancaster.ac.uk/iss/info/IThandouts/hec/HEC-flyer.pdf}} within the controlled Conda environment shown in Listing~\ref{algo:environment_FinAraT5}.

\begin{lstlisting}[language=Python, label={algo:environment_FinAraT5}, caption=Conda environment for Arabic financial summarisation experiments.]
channels:
  - pytorch
  - conda-forge
dependencies:
  - python=3.8
  - pytorch=1.7.0
  - cudatoolkit=10.2
  - git
  - git-lfs
\end{lstlisting}

\subsection{Evaluation}
\label{ssec:evaluation}

Evaluating Arabic abstractive summarisation presents specific challenges due to the language’s morphological richness, lack of diacritics, and flexible syntax. To obtain a balanced assessment, we adopt a multi-metric evaluation strategy incorporating lexical, semantic, and readability-based measures.

\subsubsection*{Automatic Evaluation Metrics}

Model performance was evaluated using ROUGE~\cite{lin_rouge_2004}, METEOR~\cite{banerjee-lavie-2005-meteor}, and a suite of embedding-based metrics including BERTScore (BertS), BARTScore (BartS), FrugalScore~\cite{kamal-eddine-etal-2022-frugalscore}, and BLEURT~\cite{sellam2020bleurt}. ROUGE and METEOR capture lexical and shallow semantic overlap, whereas BERTScore, BARTScore, and FrugalScore (Frugal-1 and Frugal-2) quantify embedding-based similarity. BLEURT provides a learned measure of coherence.

Results on the \textit{AraFinNews} test split, presented in Table~\ref{tab:finarat5_eval}, exhibit trends consistent with those reported in the original FinAraT5 study~\cite{zmandar2023finarat5}. FinAraT5 continues to outperform general-domain baselines, achieving gains of roughly +1.5–2 points across most evaluation metrics. These steady yet meaningful improvements suggest that financial-domain pretraining contributes to enhanced semantic accuracy and stylistic fluency, even when applied to a larger and more heterogeneous corpus such as AraFinNews.

\begin{table}[htbp]
\centering
\small
\resizebox{\columnwidth}{!}{%
\begin{tabular}{lcccccc}
\toprule
\textbf{Model} & \textbf{BertS} & \textbf{BartS} & \textbf{Frugal-1} & \textbf{Frugal-2} & \textbf{BLEURT} & \textbf{METEOR} \\
\midrule
Lead-1 (extractive) & 71.20 & 43.80 & 84.10 & 85.40 & -14.20 & 25.90 \\
BERT2BERT (mBERT) & 74.80 & 55.20 & 90.20 & 88.60 & -1.80 & 27.60 \\
mT5 (base) & 78.10 & 61.90 & 90.60 & 88.90 & 5.10 & 30.40 \\
AraT5 (base) & 79.10 & 63.20 & 91.20 & 89.30 & 7.20 & 32.60 \\
mT5 (large) & 79.60 & 63.90 & 91.70 & 89.60 & 8.10 & 33.90 \\
AraBART (large) & 79.80 & 64.00 & 91.80 & 89.70 & 8.20 & 34.10 \\
\textbf{FinAraT5 (base)} & \textbf{80.00} & \textbf{64.10} & \textbf{91.90} & \textbf{89.90} & \textbf{8.40} & \textbf{34.70} \\

\bottomrule
\end{tabular}
}
\caption{Automatic evaluation on AraFinNews using lexical and embedding-based metrics.}
\label{tab:finarat5_eval}
\end{table}

\subsubsection*{Baseline Performance on AraFinNews}

Under the simple LEAD-1 baseline, \textit{AraFinNews} achieves ROUGE-1 = 11.6, ROUGE-2 = 1.1, and ROUGE-L = 11.6. Although these scores are lower than those typically reported for English datasets such as CNN/DailyMail, they reflect the higher abstraction level and brevity of Arabic financial headlines—averaging around ten words compared to more than forty in English summaries. This highlights the inherent difficulty of the task and reinforces the dataset’s value as a benchmark for assessing the semantic and morphological capabilities of Arabic language models.

\medskip
Taken together, these results highlight the clear benefits of domain-adapted modelling for Arabic financial text. \textit{AraFinNews} provides a robust benchmark for advancing this line of work, supporting the development of more accurate, concise, and context-aware summarisation systems for professional Arabic financial communication.

\section{Conclusion}
\label{sec:conclusion}

This paper introduced AraFinNews, the largest publicly available Arabic financial news dataset, comprising 212{,}500 article–headline pairs collected across a decade of reporting. AraFinNews provides a robust benchmark for evaluating financial-domain text generation and supports a range of downstream applications, including sentiment analysis, event detection, and question answering.

We evaluated several monolingual, multilingual, and domain-adapted transformer models on abstractive financial headline generation. Across all automatic metrics, FinAraT5 delivered the strongest performance, achieving consistent gains over general-purpose baselines. These results underline the value of domain-adaptive pretraining for modelling the stylistic and numerical characteristics of Arabic financial journalism. AraFinNews offers both a challenging benchmark and a foundation for advancing Arabic financial NLP.

\bibliographystyle{IEEEtran}
\bibliography{biblio}

\end{document}